\documentclass{article}

\usepackage{arxiv}

\usepackage[utf8]{inputenc} 
\usepackage[T1]{fontenc}    
\usepackage{hyperref}       
\usepackage{url}            
\usepackage{booktabs}       
\usepackage{amsfonts}       
\usepackage{nicefrac}       
\usepackage{microtype}      
\usepackage{graphicx}
\usepackage{natbib}
\usepackage{doi}

\usepackage{float}
\usepackage{tikz}
\usepackage{url}

\usepackage{booktabs}  
\usepackage{subcaption}
\usepackage{amsmath}
\usepackage{amssymb}
\usepackage{amsthm}

\theoremstyle{plain}
\newtheorem{theorem}{Theorem}[section]
\newtheorem{conjecture}{Conjecture}[section]
\newtheorem{proposition}[theorem]{Proposition}

\theoremstyle{definition}
\newtheorem{definition}[theorem]{Definition}

\theoremstyle{remark}

\title{Are GNNs doomed by the topology of their input graph?}

\author{Amine Mohamed Aboussalah \\ NYU Tandon School of Engineering \\ \texttt{ama10288@nyu.edu} \\
    \And
    Abdessalam Ed-dib \\
    NYU Tandon School of Engineering \\
    \texttt{ae2842@nyu.edu}
}

\makeatletter
\def\blfootnote{\xdef\@thefnmark{}\@footnotetext}
\makeatother

\renewcommand{\headeright}{Under review -- 
 ICML2025}
\renewcommand{\shorttitle}{Are GNNs doomed by the topology of their input graph?}

\hypersetup{
pdftitle={Are GNNs doomed by the topology of their input graph?},
pdfsubject={cs.AI, cs.LG, stat.ML, math.GT},
pdfauthor={Amine M. Aboussalah, Abdessalam Ed-dib,},
pdfkeywords={Information Geometry, Graph Neural Networks, Oversmoothing},
}

\begin{document}

\maketitle

\begin{abstract}
Graph Neural Networks (GNNs) have demonstrated remarkable success in learning from graph-structured data. However, the influence of the input graph's topology on GNN behavior remains poorly understood. In this work, we explore whether GNNs are inherently limited by the structure of their input graphs, focusing on how local topological features interact with the message-passing scheme to produce global phenomena such as oversmoothing or expressive representations. We introduce the concept of $k$-hop similarity and investigate whether locally similar neighborhoods lead to consistent node representations. This interaction can result in either effective learning or inevitable oversmoothing, depending on the inherent properties of the graph. Our empirical experiments validate these insights, highlighting the practical implications of graph topology on GNN performance.
\end{abstract}


\section{Introduction}
\label{introduction}

The expressiveness of GNNs fundamentally depends on their ability to capture and transform local topological structures into meaningful representations of graph-structured data. However, although they achieved remarkable success in diverse domains ranging from molecular modeling \cite{gilmer2017neuralmessagepassingquantum} to social network analysis \cite{hamilton2018inductiverepresentationlearninglarge}, the extent to which GNNs performance is fundamentally tied to the topology of their input graphs remains poorly understood. Specifically, how does the topology of the input graph influence the transformation of local features into global representations in GNNs, and what determines whether this leads to oversmoothing \cite{chen2019measuringrelievingoversmoothingproblem} or effective, discriminative embeddings?

A key challenge lies in the interplay between two aspects: (1) the intrinsic topological structure of the input graph, which governs the connectivity patterns and influences information flow during message passing, and (2) the dynamic evolution of node representations through successive GNN layers. Although effective learning relies on exploiting these topological features, certain graph properties, such as high local similarity or poorly connected regions, can lead to phenomena like oversmoothing, where node embeddings become indistinguishable \cite{chen2019measuringrelievingoversmoothingproblem}. Understanding whether GNNs are inherently constrained by these topological factors is critical for both theoretical insights and practical advancements.


Recent work has partially addressed these issues. For example, \citet{xu2019powerfulgraphneuralnetworks} explored the expressiveness of GNNs, focusing on their ability to distinguish non-isomorphic graphs through their aggregation mechanism and readout function. In contrast, \citet{nguyen2023revisitingoversmoothingoversquashingusing} examined the link between graph curvature and oversmoothing. Meanwhile, \citet{bodnar2023neuralsheafdiffusiontopological} introduced a framework that leverages cellular sheaf theory to explain oversmoothing, modeling non-trivial sheaf structures to better understand the underlying causes of this phenomenon. 

These approaches may appear distinct, but they are fundamentally related. Whether it is curvature, the Laplacian spectrum, or cellular sheaves, they are all defined or linked through the local connectivity patterns within the input graph. Even the solutions proposed to mitigate oversmoothing rely on perturbing these connectivity patterns, such as by dropping edges or nodes, rewiring, adding residual connections, or co-training with random walks. Therefore, in this work, rather than focusing on a specific aspect of this phenomenon, we investigate its underlying source, namely its local connectivity patterns.

We examine the impact of these local connectivity patterns on the performance of GNNs through $k$-hop similarity. Here, we do not refer to $k$-hop isomorphism, where two graphs have isomorphic k-hop neighborhoods (which would be trivial). Rather, we focus on a broader case, where the sets of $k$-hop neighborhoods are identical, although the overall structure or edge arrangement may differ. 

\begin{figure}[H]
    \vskip 0.2in
    \begin{center}    \centerline{\includegraphics[width=0.5\columnwidth]{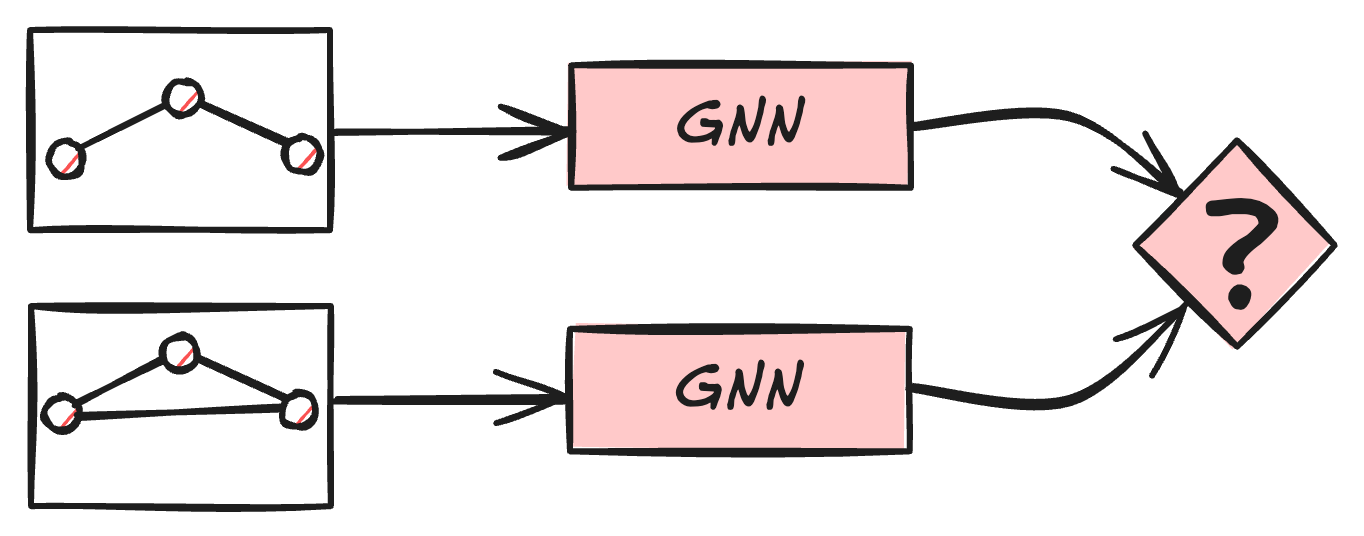}}
    \caption{Illustration of two $2$-hop similar graphs with distinct edge arrangements. Although the $2$-hop neighborhoods are identical, the graphs are not $2$-hop isomorphic.}
    \label{two-hop}
    \end{center}
    \vskip -0.2in
\end{figure}

For example, Figure \ref{two-hop} illustrates this concept: although these two graphs are not two-hop isomorphic, they are two-hop similar. The question that arises is as follows: 
\begin{center}
\textbf{\textit{Does training on two $k$-hop similar graphs lead to consistent node representations?
}}    
\end{center}

To answer this question, we propose Conjecture \ref{conjecture: weight_consistency}, which formalizes our intuition that graphs with similar local structures should induce similar learning behaviors in GNNs. Specifically, if two graphs are $k$-hop similar, the optimal parameters learned by a $k$-layer GNN should be approximately equivalent, as they are essentially solving the same local learning problems.
\begin{conjecture}[\textbf{Weight Consistency under $k$-Hop Similarity in GNNs}]
\label{conjecture: weight_consistency}
Let $G_1$ and $G_2$ be two graphs that are $k$-hop similar. Let $f_\theta$ denote a GNN with $k$ layers, and let $\theta_1$ and $\theta_2$ represent the optimal learned parameters of the GNN after training on graphs $G_1$ and $G_2$, respectively. We conjecture that the functions induced by these optimal weights, $f_{\theta_1}$ and $f_{\theta_2}$, satisfy the following approximate equality:
\begin{equation*}
    f_{\theta_1} \approx f_{\theta_2}
\end{equation*}
\end{conjecture}

In the following, we aim to validate our conjecture and explore its implications on the performance of GNNs and the oversmoothing problem. Our contributions are threefold:

\begin{itemize}
    \item We develop an algorithm to generate $k$-hop similar graphs, which will allow us to test the validity of our conjecture.
    \item We empirically validate the conjecture under various graph structures, using stochastic block models as controls.
    \item We provide an explanation for the oversmoothing phenomenon based on our conjecture, illustrating how graph topology can negatively affect GNN performance.
\end{itemize}

\section{Preliminaries}
GNNs are machine learning models designed to process graph-structured data, where entities (nodes) are connected by relationships (edges). They are widely used in applications such as social network analysis, molecular chemistry, and recommendation systems. GNNs leverage a fundamental mechanism called \textbf{message passing}, in which nodes iteratively update their representations by aggregating information from their neighbors.

The message passing process in GNNs can be viewed as a diffusion of information through the graph. At each layer $k$, the representation of a node $v$ is updated as follows:
\begin{equation*}
\mathbf{h}_v^{(k)} = \text{UPDATE}\left( \mathbf{h}_v^{(k-1)}, \text{AGGREGATE}\left( \left\{ \text{M}\left( \mathbf{h}_v^{(k-1)}, \mathbf{h}_u^{(k-1)}, \mathbf{e}_{vu} \right) : u \in \mathcal{N}(v) \right\} \right) \right)
\end{equation*}
where $ \mathbf{h}_v^{(k)} $ denotes the representation of node $ v $ at layer $ k $, $ \mathcal{N}(v) $ is the set of neighbors of node $ v $, and the aggregation function $ \text{AGGREGATE} $ combines information from the neighboring nodes $ \{ \mathbf{h}_u^{(k)} : u \in \mathcal{N}(v) \} $ according to a specified rule (e.g., sum, mean, or weighted sum).  The update rule for the node's representation, denoted by $ \text{UPDATE} $, governs how the aggregated information and the previous node representation $ \mathbf{h}_v^{(k-1)} $ are combined to produce the new representation $ \mathbf{h}_v^{(k)} $. The function $ \text{M} $ specifies how the features of node $ v $ and its neighbors interact, potentially incorporating edge features such as $ \mathbf{e}_{vu} $.

Each GNN layer corresponds to one \textit{hop} of information propagation, meaning a GNN with $k$ layers can capture information from nodes within $k$-hops of a target node, enabling it to produce expressive representations of graph features.

However, GNNs still face limitations in distinguishing certain non-isomorphic graphs. One way to construct such counterexamples is through \textbf{k-hop isomorphism}. Two graphs $ G_1 = (V_1, E_1) $ and $ G_2 = (V_2, E_2) $ are \textbf{k-hop isomorphic} if their $ k $-hop induced subgraphs are isomorphic.

\begin{definition}[\textbf{$k$-Hop Isomorphism}]
Let $ G_1 = (V_1, E_1) $ and $ G_2 = (V_2, E_2) $ be two graphs. We say that $ G_1 $ and $ G_2 $ are \textbf{k-hop isomorphic} if there exists a bijection $ f: V_1 \to V_2 $ such that for all $ v_i \in V_1 $, the induced subgraphs $ G_1[\mathcal{N}_k(v_i)] $ and $ G_2[\mathcal{N}_k(f(v_i))] $ are isomorphic, where $ \mathcal{N}_k(v) $ is the set of nodes within $ k $-hops of node $ v $.
\end{definition}

This result follows directly from the fact that the message-passing mechanism aggregates information across identical $k$-hop neighborhoods, leading to indistinguishable representations for $k$-hop isomorphic graphs.

\begin{theorem}[\textbf{Weight Consistency under $k$-Hop Isomorphism in GNNs}]
Given two graphs $ G_1 $ and $ G_2 $ that are $ k $-hop isomorphic, a GNN with $ k $ layers cannot distinguish between them. Formally, for any GNN $ f_{\theta} $ with $ k $ layers, we have:
\[
f_{\theta}(G_1) = f_{\theta}(G_2),
\]
where $ f_{\theta}(G) $ represents the learned representation of graph $ G $.
\end{theorem}

To address the strictness of $ k $-hop isomorphism, we introduce a relaxed notion called \textit{$ k $-hop similarity}, which allows for variations in edge structure of the $ k $-hop neighborhoods while preserving the set of nodes.

\begin{definition}[\textbf{$k$-Hop Similarity}]
Two graphs $ G_1 = (V_1, E_1) $ and $ G_2 = (V_2, E_2) $ are \textbf{$ k $-hop similar} if there exists a one-to-one mapping $ \phi $ between their $ k $-hop neighborhoods such that:
\[
\mathcal{N}_k(v_1) \sim \mathcal{N}_k(v_2), \quad \forall v_1 \in G_1, v_2 \in G_2,
\]
where $ \sim $ denotes an isomorphism of node sets within the $ k $-hop neighborhoods without requiring exact edge alignment.
\end{definition}

By relaxing the assumption of strict isomorphism and instead considering $k$-hop similarity, we aim to investigate whether GNNs will continue to be unable to distinguish between these graphs, as proposed in Conjecture \ref{conjecture: weight_consistency}, and how this influences their performance, particularly in relation to emerging phenomena such as oversmoothing.

In the following sections, we will first present the theoretical and empirical framework designed to validate or refute our conjecture. Based on the outcomes, we will then examine the implications of our findings for GNN performance, particularly with respect to the phenomenon of oversmoothing.

\section{Exploring $k$-Hop Similarity in GNNs}
\subsection{Definition and Properties}

As explained earlier, two graphs $G_1$ and $G_2$ are said to be $k$-hop similar if, for every node in $G_1$, there exists a corresponding node in $G_2$ such that their $k$-hop neighborhoods are identical. Here, the $k$-hop neighborhood of a node includes all nodes reachable within $k$ hops. This concept differs from $k$-hop isomorphism in that the exact connectivity between nodes within the $k$-hop neighborhoods is not required to match.

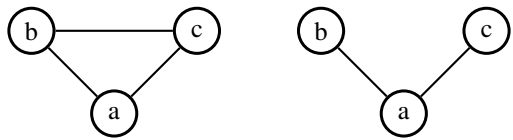
\begin{figure}[H]
\vskip 0.2in
\begin{center}
\begin{tikzpicture}[scale=1.1]

\node[circle, draw, minimum size=0.6cm, inner sep=0pt, line width=0.4mm] (a1) at (0, 1) {b};
\node[circle, draw, minimum size=0.6cm, inner sep=0pt, line width=0.4mm] (b1) at (2, 1) {c};
\node[circle, draw, minimum size=0.6cm, inner sep=0pt, line width=0.4mm] (c1) at (1, 0) {a};

\draw[thick] (a1) -- (b1);
\draw[thick] (b1) -- (c1);
\draw[thick] (c1) -- (a1);

\node[circle, draw, minimum size=0.6cm, inner sep=0pt, line width=0.4mm] (b2) at (3.5, 1) {b};
\node[circle, draw, minimum size=0.6cm, inner sep=0pt, line width=0.4mm] (a2) at (5.5, 1) {c};
\node[circle, draw, minimum size=0.6cm, inner sep=0pt, line width=0.4mm] (c2) at (4.5, 0) {a};

\draw[thick] (a2) -- (c2);
\draw[thick] (c2) -- (b2);
\end{tikzpicture}
    \end{center}
    \caption{Example of two graphs that are $2$-hop similar but not $2$-hop isomorphic. The left graph shows a complete connection between nodes $a$, $b$, and $c$, while the right graph has the same nodes connected in a path configuration.}
    \label{fig:two_hop_iso}
\end{figure}
To better understand this, consider the example in Figure \ref{fig:two_hop_iso}. Graph 1 and Graph 2 have identical 2-hop neighborhoods for their respective nodes; however, the induced subgraphs (consisting of nodes within 2 hops and the edges between them) for node ``$a$'' are not isomorphic\footnote{The left graph will induce a complete connection between nodes $a$, $b$, and $c$, while the right graph has the same nodes connected in a path configuration.}. Indeed, $k$-hop isomorphism implies $k$-hop similarity, as the latter requires only matching $k$-hop neighborhoods without enforcing identical edge connectivity within these neighborhoods.

\begin{proposition}[\textbf{Isomorphism Implies Similarity}] If two graphs are $k$-hop isomorphic, they are necessarily $k$-hop similar, but the reverse is not always true. \end{proposition}

An alternative interpretation of $k$-hop similarity is that $G_1$ and $G_2$ are considered $k$-hop similar if they can be viewed as $k$-th roots of the same graph $G$. In other terms, $G$ is the $k$-th power graph of both graphs $G1$ and $G_2$.

\begin{definition}[\textbf{$k$-th Power of a Graph}] 
The $k$-th power of a graph $G = (V, E)$, denoted $G^k$, is a graph where there is an edge between nodes $u$ and $v$ if and only if the shortest path distance between them in $G$ is at most $k$. 
\end{definition}

This leads to a practical test for $k$-hop similarity: $G_1$ and $G_2$ are $k$-hop similar if and only if their $k$-th power graphs are identical.

\begin{proposition}[\textbf{Binary $k$-Hop Reachability}]Let $\mathbf{A}_1$ and $\mathbf{A}_2$ be the adjacency matrices of $G_1$ and $G_2$, respectively. Then $G_1$ and $G_2$ are $k$-hop similar if and only if their binary $k$-hop reachability matrices satisfy:
$$R_1 = R_2,$$
where the binary $k$-hop reachability matrix $R_i$ is defined as $R_i = (A_1 + A_i^2 + \cdots + A_i^k > 0)$.
\end{proposition}

Now, the key question arises: how can we generate $k$-hop similar graphs, as we will need them to validate our hypothesis? Intuitively, one might consider computing the $k$-th power graph and then finding its roots. However, this approach is computationally challenging.

\begin{theorem}[\textbf{NP-Hardness}] Computing the roots of the $k$-th power graph is NP-hard.\end{theorem}

\subsection{Constructing $k$-Hop Similar Graphs}
Rather than directly computing the roots of the $k$-th power graph for a given input graph, we propose perturbing the original graph by removing edges, while preserving $k$-hop similarity. To achieve this, we focus on non-critical edges, which we define as edges that do not impact the $k$-hop neighborhoods of the graph.

Specifically, we aim to remove existing edges in such a way that the $k$-hop neighborhoods remain unchanged. To verify that the $k$-hop similarity is preserved, we use the distance matrix, which provides the shortest path distance between any two nodes. If two nodes are within a distance of $k$, we ensure that the operation does not increase the distance beyond $k$. Similarly, if the distance between two nodes is greater than $k$, our operation should not reduce it to less than $k$. To compute the distance matrix, we employ the Floyd-Warshall algorithm.  Algorithm \ref{alg:khopbasic} summarizes our procedure.
\begin{algorithm}[H]
\caption{Generate $K$-Hop Similar Graph}
\label{alg:khopbasic}
\begin{algorithmic}[1]
\REQUIRE{Adjacency matrix $A$, hop count $k$}
\ENSURE{A $k$-hop similar graph $G_{k}$}

\STATE $E \gets \textsc{GetEdges}(A)$
\STATE $G_{k} \gets A$ \hfill $\triangleright$ Copy of adjacency matrix
\STATE $D \gets \textsc{FloydWarshall}(G_{k})$
\STATE $paths_{k} \gets [D \leq k]$ \hfill $\triangleright$ $k$-hop paths

\FOR{each edge $(i,j)$ in $E$}
    \STATE $G_{temp} \gets G_{k}$ \hfill Temporary copy of current graph
    \STATE $G_{temp}[i,j] \gets 0$
    \STATE $G_{temp}[j,i] \gets 0$
    
    \STATE $D_{temp} \gets \textsc{FloydWarshall}(G_{temp})$
    \STATE $paths_{temp} \gets  [D_{temp} \leq k]$
    
    \IF{$paths_{temp} = paths_{k}$}
        \STATE $G_{k}[i,j] \gets 0$
        \STATE $G_{k}[j,i] \gets 0$
        \STATE $D \gets D_{temp}$
    \ENDIF
    
\ENDFOR

\STATE \textbf{return} $G_{k}$
\end{algorithmic}
\end{algorithm}

However, the complexity of this algorithm is quite high, which makes it impractical for very large graphs. The complexity arises from the need to traverse the edge set $E$ for edge removal, and for each edge traversal, compute the distance matrix using Floyd-Warshall \cite{floyd_warshall}, which has a time complexity of $O(V^3)$, where $V$ is the number of vertices. Thus, resulting in an overall complexity of $O(E \times V^3)$.

To address this high complexity, we propose the following modifications:

\begin{enumerate}
    \item \textbf{Threshold on Removals}: We introduce a threshold on the number of edges to remove. The goal is to ensure that the generated graph is structurally different from the original graph. A common measure of structural difference is the graph edit distance, which counts the minimum number of operations required to transform one graph into another. In our case, this would correspond to the number of edges we removed. We define the threshold on the graph edit distance, and once this threshold is reached, we stop exploring further edge modifications.

    \item \textbf{Batching Edge Modifications}: Instead of modifying one edge at a time, we can perform edge removals in batches. This reduces the number of edge traversals and the number of times the distance matrix needs to be recomputed. By processing multiple edges in a single step, we can significantly reduce the overall computational complexity.
\end{enumerate}

\begin{algorithm}[H]
\caption{Generate K-Hop Similar Graph with Batch Processing and Removal Thresholding}
\label{alg:improved_khop}
\begin{algorithmic}[1]
\REQUIRE{Adjacency matrix $A$, hop count $k$, removal threshold $T$, batch size $b$}
\ENSURE{A $k$-hop similar graph $G_k$}
\STATE $(E \gets \textsc{GetEdges}(A)$ \hfill $\triangleright$ Get edge and complementary lists
\STATE $G_{k} \gets A$ \hfill $\triangleright$ Copy of adjacency matrix
\STATE $D \gets \textsc{FloydWarshall}(G_{k})$
\STATE $path_{k} \gets [D\leq k]$ \hfill $\triangleright$$k$-hop paths
\STATE $removals \gets 0$
\FOR{$start \gets 0$ to $|E|$ step $b$}
    \STATE $batch \gets E[start:start+b]$ \hfill $\triangleright$Get current batch of edges
    \STATE $G_{temp} \gets G_{k}$ \hfill $\triangleright$ Temporary copy of current graph
    \FOR{each edge $(i,j)$ in $batch$}
        \STATE $G_{temp}[i,j] \gets 0$
        \STATE $G_{temp}[j,i] \gets 0$
    \ENDFOR  
    \STATE $D_{temp} \gets \textsc{FloydWarshall}(G_{temp})$
    \STATE $path_{temp} \gets [D_{temp} \leq k]$
    \IF{$path_{temp} = path_{k}$}
        \FOR{each edge $(i,j)$ in $batch$}
            \STATE $G_{k}[i,j] \gets 0$
            \STATE $G_{k}[j,i] \gets 0$
        \ENDFOR
        \STATE $D \gets D_{temp}$
        \STATE $removals \gets removals + |batch|$
    \ENDIF   
    \IF{$removals \geq T$}
        \STATE \textbf{return} $G_{k}$
    \ENDIF
\ENDFOR
\STATE \textbf{return} $G_{k}$
\end{algorithmic}
\end{algorithm}
By introducing these modifications, we aim to reduce the time complexity of Algorithm \ref{alg:khopbasic} while maintaining the core objective of perturbing the graph without violating $k$-hop similarity. The threshold on removals controls structural changes by limiting edge modifications, while batching improves efficiency by grouping operations, thus reducing the number of distance matrix computations. The resulting complexity of $O\left(\frac{T \times V^3}{b}\right)$ significantly improves efficiency for larger graphs, where $T$ is the number of edges to modify, $b$ is the batch size and $V$ is the number of nodes. Algorithm \ref{alg:improved_khop} provides a  summary of our approach for constructing $k$-hop similar graphs.

\subsection{Conjecture Validation}
In this section, we conduct a series of experiments to validate our conjecture. To evaluate it across different graph structures, we focus on node classification using synthetic datasets generated with Stochastic Block Models (SBMs) \cite{stochastic_block_models}. SBMs provide a controlled environment for analyzing the influence of structural properties and closely resemble community-structured real-world networks, making them a suitable benchmark for studying the impact of graph structure on our conjecture validity.

After training the GNNs, we validate our conjecture by measuring the level of disagreement between the learned functions trained on two $k$-hop similar graphs. By quantifying disagreement, we aim to understand the sensitivity of learned representations to local structural changes. 

In this context, we define the level of disagreement as: 
\begin{definition}[\textbf{Level of Disagreement}]
The level of disagreement measures the fraction of nodes where the predictions differ between the original graph $ G_1 $ and the $k$-hop similar graph $ G_2 $. Formally:
\[
\text{Disagreement} = \frac{1}{N} \sum_{i=1}^N \mathbb{I}(\hat{y}_i^{(1)} \neq \hat{y}_i^{(2)}),
\]
where:
\begin{itemize}
    \item $ \hat{y}_i^{(1)} $ and $ \hat{y}_i^{(2)} $ are the predicted labels for node $ i $ in $ G_1 $ and $ G_2 $, respectively,
    \item $ \mathbb{I}(\cdot) $ is the indicator function,
    \item $ N $ is the total number of nodes.
\end{itemize}
\end{definition}
\subsubsection{Synthetic Dataset Generation}
We conduct node classification experiments on synthetic graphs generated using stochastic block models (SBMs) to analyze the sensitivity of our assumptions to variations in graph structure. Our base configuration consists of $1500$ nodes, intracluster edge probability of $0.5$, intercluster edge probability of $0.1$, and $2$ classes. To isolate the influence of each parameter, we perform controlled experiments where we fix all other parameters and vary one at a time. Specifically, we vary the number of nodes between $50$ and $3000$, intra-cluster edge probability between $0.1$ and $0.9$, inter-cluster edge probability between $0.1$ and $0.9$, and the number of classes between $2$ and $10$.

Node features are sampled from a $1024$-dimensional Gaussian distribution, where each class has a unique mean vector equal to its class label, and variance is set to $1$. The dataset is approximately balanced, with an equal number of nodes per class. We use a $80/20/20$ split for training, validation, and testing, respectively
 
\subsubsection{Experimental Setting}
For $k$-hop similar graph generation, we use Algorithm \ref{alg:improved_khop} and set the graph edit distance threshold to $20\%$ of the edges in the original input graph. The batch size is chosen such that it corresponds to half of the edges to be removed. This ensures a trade-off between generating diverse graph structures while maintaining $k$-hop similarity to the original graph.

We train a two-layer Graph Convolutional Neural Network (GCN) \cite{kipf2017semisupervisedclassificationgraphconvolutional} with ReLU activation, and $32$ hidden units.

All models are trained using the Adam optimizer \cite{kingma2017adammethodstochasticoptimization}  with a learning rate of 0.01 for up to $200$ epochs. Early stopping is applied based on validation loss, using a window size of $50$ epochs. Performance is evaluated using accuracy and the level of disagreement across all datasets. 

All experimental results are averaged over $10$ independent runs with different random initializations to ensure statistical robustness, reporting both mean and standard deviation (mean $\pm$ std).

\subsubsection{Numerical Results}
\paragraph{Number of nodes.}
Table \ref{tab:number_nodes} compares the performance of the model on original and $k$-hop similar graphs with varying node counts (ranging from $50$ to $3,000$). The results strongly validate the weight consistency conjecture, showing that $ f_{\theta_1} \approx f_{\theta_2} $ holds for different graph scales. This is reflected in the exceptionally low disagreement rates ($\leq 0.69\%$) and nearly identical accuracy scores between graph pairs, despite notable structural differences ($20\%$ edges are different).
\begin{table}[H]
\caption{Performance comparison across different number of nodes. Results show classification accuracy (\%) on original and $K$-hop similar graphs, and level of disagreement (\%). All metrics are reported as mean $\pm$ standard deviation over $10$ runs.}
\label{tab:number_nodes}
\vskip 0.15in
\begin{center}
\resizebox{0.6\columnwidth}{!}{
\begin{small}
\begin{sc}
\begin{tabular}{r|ccc}
\toprule
\textbf{Nodes} & \textbf{Original Graph Acc.} & \textbf{K-Hop Graph Acc.} & \textbf{Disagreement} \\
\midrule
50 & 94.00 $\pm$ 18.00 & 94.00 $\pm$ 18.00 & 0.20 $\pm$ 0.60  \\
100 & 100.00 $\pm$ 0.00 & 100.00 $\pm$ 0.00 & 0.00 $\pm$ 0.00  \\
500 & 92.40 $\pm$ 3.35 & 92.20 $\pm$ 2.56 & 0.32 $\pm$ 0.39  \\
1000 & 98.45 $\pm$ 2.81 & 97.75 $\pm$ 3.24 & 0.69 $\pm$ 1.06 \\
1500 & 99.97 $\pm$ 0.10 & 99.97 $\pm$ 0.10 & 0.06 $\pm$ 0.18 \\
2000 & 98.65 $\pm$ 2.64 & 98.67 $\pm$ 2.65 & 0.02 $\pm$ 0.03  \\
2500 & 99.26 $\pm$ 2.22 & 99.26 $\pm$ 2.22 & 0.00 $\pm$ 0.00  \\
3000 & 100.00 $\pm$ 0.00 & 100.00 $\pm$ 0.00 & 0.00 $\pm$ 0.00  \\
\bottomrule
\end{tabular}
\end{sc}
\end{small}
}
\end{center}
\vskip -0.1in
\end{table}

\paragraph{Intracluster Probability.}

Table \ref{tab:intraclass_performance} explores the effect of varying intra-class connection probabilities (ranging from $0.2$ to $0.9$) on the consistency of the learned functions. The results strongly support $ f_{\theta_1} \approx f_{\theta_2} $, showing extremely low disagreement rates ($\leq 0.25\%$) and negligible differences in accuracy (maximum of $0.03\%$) between original and $k$-hop similar graphs. This consistency holds across all values of intraclass probability, implying that the learned functions remain largely unchanged regardless of the density of connections within the same class.
\begin{table}[H]
\caption{Performance comparison across different intraclass probability values. Results show classification accuracy (\%) on original and $K$-hop similar graphs, and level of disagreement (\%). All metrics are reported as mean $\pm$ standard deviation over $10$ runs.}
\label{tab:intraclass_performance}
\vskip 0.15in
\begin{center}
\resizebox{0.6\columnwidth}{!}{
\begin{small}
\begin{sc}
\begin{tabular}{r|ccc}
\toprule
\textbf{Intra-Class} & \textbf{Original Graph Acc.} & \textbf{K-Hop Graph Acc.} & \textbf{Disagreement} \\
\textbf{Probability} & \textbf{(\%)} & \textbf{(\%)} & \textbf{(\%)} \\
\midrule
0.2 & 88.77 $\pm$ 1.54 & 88.77 $\pm$ 1.54 & 0.00 $\pm$ 0.00 \\
0.3 & 90.80 $\pm$ 8.98 & 90.77 $\pm$ 9.07 & 0.25 $\pm$ 0.34 \\
0.4 & 99.50 $\pm$ 1.50 & 99.53 $\pm$ 1.40 & 0.03 $\pm$ 0.08 \\
0.5 & 99.40 $\pm$ 1.80 & 99.43 $\pm$ 1.70 & 0.05 $\pm$ 0.14 \\
0.6 & 97.77 $\pm$ 3.48 & 97.73 $\pm$ 3.51 & 0.02 $\pm$ 0.03 \\
0.7 & 98.97 $\pm$ 2.07 & 99.00 $\pm$ 2.01 & 0.07 $\pm$ 0.22 \\
0.8 & 100.00 $\pm$ 0.00 & 100.00 $\pm$ 0.00 & 0.00 $\pm$ 0.00 \\
0.9 & 100.00 $\pm$ 0.00 & 100.00 $\pm$ 0.00 & 0.00 $\pm$ 0.00 \\
\bottomrule
\end{tabular}
\end{sc}
\end{small}
}
\end{center}
\vskip -0.1in
\end{table}

\paragraph{Intercluster Probability.}
Table \ref{tab:interclass_performance} investigates the impact of varying interclass connection probabilities (from $0$ to $0.9$) on function consistency. Although the overall accuracy decreases with higher interclass connectivity, the approximation $ f_{\theta_1} \approx f_{\theta_2} $ remains strong, as evidenced by low disagreement rates ($\leq 0.51\%$) and consistent accuracy scores (maximum difference of $0.32\%$). This suggests that the weight consistency conjecture continues to hold even as the graph structure becomes more complex with increased interclass connectivity.
\begin{table}[H]
\caption{Performance comparison across different interclass probability values. Results show classification accuracy (\%) on original and $K$-hop similar graphs, and level of disagreement (\%). All metrics are reported as mean $\pm$ standard deviation over $10$ runs.}
\label{tab:interclass_performance}
\vskip 0.15in
\begin{center}
\resizebox{0.6\columnwidth}{!}{
\begin{small}
\begin{sc}
\begin{tabular}{r|ccc}
\toprule
\textbf{Inter-Class} & \textbf{Original Graph Acc.} & \textbf{K-Hop Graph Acc.} & \textbf{Disagreement} \\
\textbf{Probability} & \textbf{(\%)} & \textbf{(\%)} & \textbf{(\%)} \\
\midrule
0.0 & 100.00 $\pm$ 0.00 & 100.00 $\pm$ 0.00 & 0.00 $\pm$ 0.00 \\
0.1 & 99.60 $\pm$ 1.60 & 99.92 $\pm$ 0.30 & 0.28 $\pm$ 1.12 \\
0.2 & 99.73 $\pm$ 1.32 & 99.94 $\pm$ 0.24 & 0.20 $\pm$ 0.92 \\
0.3 & 92.57 $\pm$ 13.23 & 92.47 $\pm$ 13.67 & 0.51 $\pm$ 1.22 \\
0.4 & 86.91 $\pm$ 16.47 & 86.83 $\pm$ 16.73 & 0.41 $\pm$ 1.11 \\
0.5 & 83.68 $\pm$ 16.70 & 83.61 $\pm$ 16.90 & 0.34 $\pm$ 1.02 \\
0.6 & 80.72 $\pm$ 17.11 & 80.66 $\pm$ 17.27 & 0.30 $\pm$ 0.96 \\
0.7 & 78.55 $\pm$ 17.04 & 78.49 $\pm$ 17.17 & 0.26 $\pm$ 0.90 \\
0.8 & 76.87 $\pm$ 16.78 & 76.83 $\pm$ 16.89 & 0.23 $\pm$ 0.85 \\
0.9 & 75.66 $\pm$ 16.35 & 75.62 $\pm$ 16.46 & 0.21 $\pm$ 0.81 \\
\bottomrule
\end{tabular}
\end{sc}
\end{small}
}
\end{center}
\end{table}

\paragraph{Number of classes.}
Table \ref{tab:classes_performance} reveals a surprising behavior when examining the weight consistency conjecture across different numbers of classes (ranging from $2$ to $10$). While \( f_{\theta_1} \approx f_{\theta_2} \) holds strongly for binary and three-class problems (with disagreement \(\leq 4.24\%\)), we observe an unexpected sharp increase in disagreement rates for higher numbers of classes, reaching up to $81.08\%$ for $10$ classes. Intriguingly, despite this high disagreement in individual predictions, both models maintain remarkably similar overall accuracy scores (with a maximum difference of $2.47\%$).
\begin{table}[H]
\caption{Performance comparison across different numbers of classes. Results show classification accuracy (\%) on original and $K$-hop similar graphs, and level of disagreement (\%). All metrics are reported as mean $\pm$ standard deviation over $10$ runs.}
\label{tab:classes_performance}
\vskip 0.15in
\begin{center}
\resizebox{0.6\columnwidth}{!}{
\begin{small}
\begin{sc}
\begin{tabular}{r|ccc}
\toprule
\textbf{Number of} & \textbf{Original Graph Acc.} & \textbf{K-Hop Graph Acc.} & \textbf{Disagreement} \\
\textbf{Classes} & \textbf{(\%)} & \textbf{(\%)} & \textbf{(\%)} \\
\midrule
2 & 99.43 $\pm$ 0.86 & 99.00 $\pm$ 1.48 & 0.63 $\pm$ 1.05 \\
3 & 78.27 $\pm$ 7.83 & 77.30 $\pm$ 6.54 & 4.24 $\pm$ 4.54 \\
4 & 72.20 $\pm$ 6.07 & 70.10 $\pm$ 3.49 & 4.30 $\pm$ 4.83 \\
5 & 44.10 $\pm$ 6.05 & 46.57 $\pm$ 9.35 & 45.95 $\pm$ 33.28 \\
6 & 28.87 $\pm$ 8.25 & 28.67 $\pm$ 7.70 & 73.81 $\pm$ 30.42 \\
7 & 29.37 $\pm$ 7.82 & 29.83 $\pm$ 8.11 & 54.43 $\pm$ 35.84 \\
8 & 19.63 $\pm$ 5.01 & 20.30 $\pm$ 4.90 & 78.47 $\pm$ 16.74 \\
9 & 21.17 $\pm$ 2.71 & 21.23 $\pm$ 3.89 & 77.79 $\pm$ 6.88 \\
10 & 17.33 $\pm$ 4.67 & 17.60 $\pm$ 4.82 & 81.08 $\pm$ 20.75 \\
\bottomrule
\end{tabular}
\end{sc}
\end{small}
}
\end{center}
\end{table}
Investigating the predicted class probabilities reveals that this apparent contradiction arises from model uncertainty. As the number of classes increases, the models produce increasingly uniform probability distributions across classes, indicating low confidence in their predictions. This explains why small perturbations in input graphs can lead to different class assignments while maintaining similar accuracy levels: essentially, the models are making near-random choices among equally uncertain options. In other words, while the models may assign different classes due to subtle changes in the input, the lack of clear discrimination between classes results in similar accuracy outcomes. 

\begin{figure}[H]
    \vskip 0.2in
    \begin{center}    \centerline{\includegraphics[width=0.6\columnwidth]{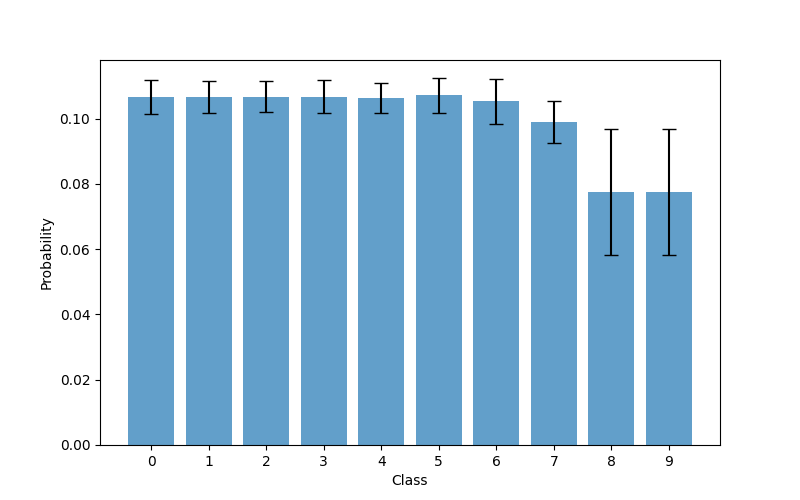}}
    \caption{Average predicted probability distribution across ten classes for a GNN model trained on the original graph.}
    \label{origin}
    \end{center}
    \vskip -0.2in
\end{figure}

\begin{figure}[H]
    \vskip 0.2in
    \begin{center}
    \centerline{\includegraphics[width=0.6\columnwidth]{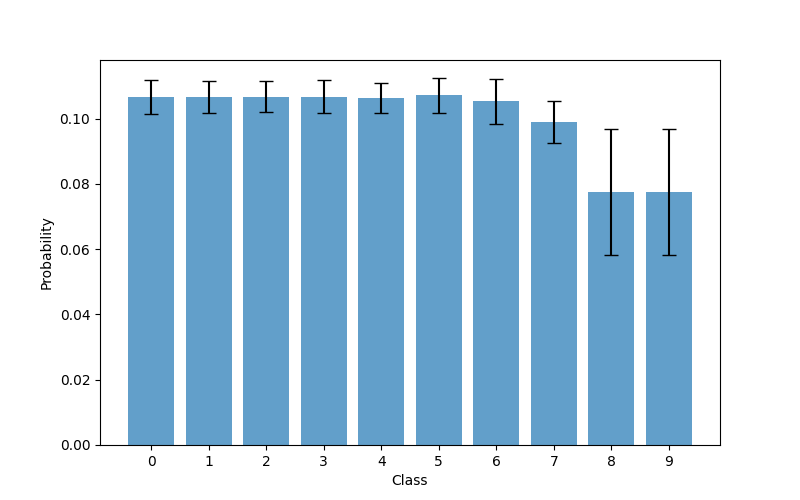}}
    \caption{Average predicted probability distribution across ten classes for a GNN model trained on the $2$-hop similar graph.}
    \label{hophop}
    \end{center}
    \vskip -0.2in
\end{figure}

Figures \ref{origin} and \ref{hophop} present the average predicted probability distribution for disagreed nodes across ten classes, comparing the models trained on the original graph and the two-hop similar graph. The near-uniform distribution (with probabilities ranging from approximately 0.08 to 0.10) indicates the model's low confidence in class assignments. Each class receives similar probability mass, regardless of the true label, which suggests that the model is uncertain in distinguishing between the classes.

This behavior suggests that the issue is not with the weight consistency conjecture itself, but rather with the model's fundamental inability to learn discriminative features for multi-class classification tasks.

Based on our experimental analysis, we consider that Conjecture \ref{conjecture: weight_consistency} is \textbf{valid}.

\section{Oversmoothing through Graph Topology}
In the previous section, we empirically established that training the same GNN model architecture on a graph and its $k$-hop similar graph results in consistent predictions (Conjecture \ref{conjecture: weight_consistency}). In this section, we explore how this helps explain the oversmoothing phenomenon in deep GNNs.

To begin, consider a connected graph. For a sufficiently large value of $k$, the $k$-th power of this graph becomes a complete graph.

\begin{proposition}[\textbf{$k$-th Power of a Connected Graph}] The $k$-th power of a connected graph is a complete graph. \end{proposition}

This means that when training a very deep GNN on a connected graph, the effect of training is equivalent to training it on a complete graph. In the context of the GNN message-passing scheme, training on a complete graph causes all node features to be mixed together. As a result, node representations converge to identical values, which is the essence of oversmoothing. This can be formally expressed as:
\begin{theorem}[\textbf{$|V|$-hop Similarity}]
\label{thm: vhop}
Let $f_\theta$ denote a deep GNN parameterized by $\theta$, and let $G = (V, E)$ be a connected graph, where $V$ is the set of vertices and $E$ is the set of edges. Suppose $f_\theta$ is trained on graph $G$. We conjecture that training $f_\theta$ on $G$ is approximately equivalent to training it on the complete graph $K_{|V|}$, provided that the depth of the GNN is sufficiently large. Formally, if $f_\theta^{G}$ denotes the function induced by the GNN trained on graph $G$, and $f_\theta^{K_{|V|}}$ denotes the function induced by training the GNN on the complete graph, then for a sufficiently deep GNN: $f_\theta^{G} \approx f_\theta^{K_{|V|}}.$
\end{theorem}

This explains why deep GNNs often suffer from oversmoothing, where node features become indistinguishable as the network depth increases.

In practical settings, however, many real-world graphs are not fully connected; instead, they exhibit community structures with distinct connected components. In these cases, oversmoothing will occur within each connected component or cluster. Specifically, Theorem \ref{thm: vhop} can be generalized to arbitrary graph structures, demonstrating that the equivalence property holds independently within each connected component. This is formalized in Theorem \ref{thm:compo_hop}, which explains how oversmoothing manifests within each cluster.

\begin{theorem}[\textbf{Component-wise $|V_i|$-hop Similarity}]
\label{thm:compo_hop}
Let $f_\theta$ denote a deep GNN parameterized by $\theta$, and let $G = (V, E)$ be a graph with k connected components ${G_i = (V_i, E_i)}_{i=1}^k$, where $\bigcup_{i=1}^k V_i = V$ and $\bigcup_{i=1}^k E_i = E$. Suppose $f_\theta$ is trained on graph $G$. We conjecture that training $f_\theta$ on $G$ is approximately equivalent to training it on a graph $K_G$ that is the union of complete graphs, where each complete graph corresponds to a connected component of $G$. Formally, let $K_G = \bigcup_{i=1}^k K_{|V_i|}$, where $K_{|V_i|}$ is the complete graph on vertex set $V_i$. If $f_\theta^G$ denotes the function induced by the GNN trained on graph $G$, and $f_\theta^{K_G}$ denotes the function induced by training the GNN on $K_G$, then for a sufficiently deep GNN: $f_\theta^G \approx f_\theta^{K_G}$.
\end{theorem}

This suggests that, in graphs with a community structure or multiple disconnected components, oversmoothing may not be a global issue but could manifest separately within each community. As a result, even shallow GNNs operating on these graphs might suffer from oversmoothing within individual components, which can undermine the ability of the model to capture meaningful distinctions between nodes. In other words, \textbf{\textit{GNNs that rely on the message-passing scheme are fundamentally constrained by the topology of their input graphs}}, which dictates both their expressiveness and their susceptibility to oversmoothing.
\section{Conclusion and Future Work}
In this work, we investigated the influence of graph topology on the behavior of GNNs, with a focus on how local topological features, such as $k$-hop similarity, interact with the message-passing scheme to impact global learning dynamics. Our theoretical framework, grounded in the concept of $k$-hop similarity, provides new insights into how locally similar neighborhoods induce consistent learning patterns across graphs. This interaction leads to either effective learning or inevitable oversmoothing, depending on the graph's inherent properties. Through empirical validation, we demonstrated that $k$-hop similar graphs lead to consistent GNN predictions, reinforcing our conjecture of weight consistency under $k$-hop similarity. Furthermore, we offered a topological explanation for oversmoothing, showing how deep GNNs trained on graphs with large $k$ values effectively transform each connected component into a complete graph, resulting in node representations that converge to identical values within each component. This sheds light on why oversmoothing occurs in deep GNNs but also in shallow ones, particularly in graphs with community structures or disconnected components.

Despite these insights, our framework still faces certain limitations, particularly in the scalability of our $k$-hop similar graph generation algorithm and the comprehensiveness of our empirical validation. Future work should focus on developing alternative algorithms that can efficiently handle large graphs. Additionally, more extensive empirical validation is crucial. This includes conducting ablation studies to assess the impact of different $k$ values, evaluating performance across various downstream tasks and GNN architectures, and testing on real-world datasets to further substantiate our findings.

\bibliographystyle{unsrtnat}
\bibliography{references} 

\end{document}